\PassOptionsToPackage{numbers}{natbib}

\documentclass[11pt, a4paper]{deepmind}
\usepackage{kantlipsum, lipsum}
\usepackage{dm-colors}

\usepackage[utf8]{inputenc} 
\usepackage[T1]{fontenc}    
\usepackage{hyperref}       
\usepackage{url}            
\usepackage{booktabs}       
\usepackage{amsfonts}       
\usepackage{nicefrac}       
\usepackage{microtype}      
\usepackage{caption}
\usepackage{subcaption}
\usepackage[title]{appendix}
\usepackage{amsmath}
\usepackage{graphicx}
\usepackage{cleveref}
\usepackage{xcolor}
\usepackage{multirow}
\usepackage{bm}
\usepackage[subpreambles=true]{standalone}

\newcommand*{\defeq}{\mathrel{\vcenter{\baselineskip0.5ex \lineskiplimit0pt
                     \hbox{\scriptsize.}\hbox{\scriptsize.}}}%
                     =}

\DeclareMathOperator*{\expec}{\mathbb{E}}

\correspondingauthor{polc@google.com, theophane@google.com}
\title{Neural Recursive Belief States in Multi-Agent Reinforcement Learning}

\author[ \hspace{-1ex}]{Pol Moreno}
\author[ \hspace{-1ex}]{Edward Hughes}
\author[ \hspace{-1ex}]{Kevin R. McKee}
\author[ \hspace{-1ex}]{Bernardo Avila Pires}
\author[ \hspace{-1ex}]{Th\'eophane Weber}

\affil[ \hspace{-1ex}]{DeepMind}

\renewcommand{\today}{} 

\begin{abstract}
    In multi-agent reinforcement learning, the problem of learning to act is particularly difficult because the policies of co-players may be heavily conditioned on information only observed by them. On the other hand, humans readily form beliefs about the knowledge possessed by their peers and leverage beliefs to inform decision-making. Such abilities underlie individual success in a wide range of Markov games, from bluffing in Poker to conditional cooperation in the Prisoner's Dilemma, to convention-building in Bridge. Classical methods are usually not applicable to complex domains due to the intractable nature of hierarchical beliefs (i.e.\ beliefs of other agents' beliefs). We propose a scalable method to approximate these belief structures using recursive deep generative models, and to use the belief models to obtain representations useful to acting in complex tasks. Our agents trained with belief models outperform model-free baselines with equivalent representational capacity using common training paradigms. We also show that higher-order belief models outperform agents with lower-order models.
\end{abstract}

\begin{document}
\maketitle
\section{Introduction}

Partially observable Markov games \cite{Littman:1994:MGF:3091574.3091594, Hansen:2004:DPP:1597148.1597262} (POMGs) are the most general and most challenging category of learning problems for multi-agent systems. Here, agents may have competing goals, the full state of the world is not observable, and the environment appears non-stationary to any individual, by virtue of the learning dynamics of their co-players.\footnote{In a sense, multi-agent problems are generically partially observable, since the policy distributions of each player are not usually common knowledge at each point in time, whereas sampled actions may be observed.}  Multi-agent reinforcement learning (MARL) \cite{4445757} provides a method for learning capable policies in POMGs. However, the equilibria found by model-free systems are not necessarily optimal from the perspective of the individual or the collective. Different types of POMGs present unique failure modes. In competitive settings, agents may learn to play an unexploitable strategy, rather than adapting to the weaknesses of their opponents for greater payoff, as discussed in \cite{vezhnevets2019options}. If the environment is a social dilemma, MARL often converges to defecting equilibria, motivated by greed or fear, see for example \cite{DBLP:journals/corr/abs-1803-08884, DBLP:journals/corr/LererP17}. Coordination problems typically have multiple equilibria, and selecting the best is an extremely hard joint exploration problem; see \cite{Claus:1998:DRL:295240.295800} for instance. To address these difficulties, we can leverage a belief model over the unobserved parts of the world, comprising the private information of other players and their policies. This is often referred to as opponent modelling \cite{DBLP:journals/ai/AlbrechtS18}, although it may be applied outside of purely competitive situations. In the human behavioral literature, this ability is known as theory of mind \cite{premack_woodruff_1978, doi:10.1080/17405629.2018.1435413}. Prior work has shown success in the tabular setting by learning belief models with frequentist \cite{Chang2001PlayingIB, Makino:2006:MRL:1160633.1160772} and Bayesian \cite{Chalkiadakis:2003:CMR:860575.860689} statistical estimation. However, scaling these approaches beyond small tabular settings is notoriously difficult \cite{doshi}.

With the advent of deep reinforcement learning (RL), it has become tractable to learn embeddings which represent the policies of others \cite{rabinowitz2018machine}. However, combining such learning with an adaptive rule which simultaneously leverages the model remains a difficult challenge. In many settings, fortunately, we have access to interpretable representations of the world that are valuable for the task at hand; an example of such is the private information of others. Indeed, the mental states that humans attribute to others are often characterized as representing private information about the world \cite{byom}. We leverage a learned predictive model of such representations as input to the policy. Crucially, we do not otherwise alter the policy update, allowing us to retain the scalability and domain-agnosticism of deep RL. Furthermore, our method includes the recursive reasoning \cite{stahl1995players} characteristic of humans, by learning models over the belief states of other players. 

The method we propose learns beliefs using recursive deep generative models, which are designed to provide useful representations for acting in multi-agent tasks.
In our experiments, we demonstrate that agents equipped with these belief models learn to leverage the behavior of their co-players to maximize rewards in an intricate partially observable gridworld domain (Running with Scissors \cite{vezhnevets2019options}).

\section{Methods}
\subsection{Partially observable Markov games}
\label{sec:belief_models}

Partially Observable Markov Decision Processes (POMDPs) provide a general framework to study the behavior of a solitary, rational agent interacting with an environment whose true state $S_t$ is unknown to the agent. At every time-step, the agent receives sensory observation $Y_t$, sampled from the observation distribution $p(Y_t|S_t)$. The agent uses all the information known, the \textit{history} of all past observations $H_t = (Y_{\leq t}, A_{\leq t-1})$, to take an action $A_t$ sampled from the policy $\pi(A_t|H_t)$. The agent receives a reward $R_t(S_t,A_t)$ and the environment state changes according to its transition dynamics $p(S_{t+1}|S_t,A_t)$. The aim of the agent is to maximize the expected sum of discounted rewards. A particular challenge in POMDPs is the growth of relevant information available to the agent over time: it is not clear how to maintain a concise representation of knowledge which is sufficient to compute optimal decisions. Given an appropriate prior, the observation distribution and transition dynamics, one can compute by Bayes' rule the posterior distribution $p(S_t|H_t)$ over the true state of the environment given all information available at time $t$. A key result \cite{astrom1965optimal} states that this is sufficient to compute an optimal policy; this quantity is called the \emph{belief state} (or belief for short) and represents the knowledge an agent has over the true state of the world. Classically, belief states were represented by simple distributions and computed by explicit Bayesian filtering \cite{ross2011bayesian}, which limited their applicability to complex environments. However, a recent body of work
has been leveraging deep learning techniques to represent a belief state by a neural code $\hat{b}_t$. This code is computed by a recurrent neural network (RNN) $\hat{b}_t\defeq\text{RNN}(\hat{b}_{t-1}, Y_t)$, and follows an update mimicking the Bayesian filter. This code can be decoded to a state distribution by a deep generative model $p(S_t|\hat{b}_t)$. These \emph{neural belief states} are either learned using supervision on state information \cite{moreno2018neural,humplik2019meta}, or learned indirectly through unsupervised learning \cite{venkatraman2017predictive,IglZLWW18,GregorPBBW19,guo2018neural,zhang2019learning,gregor2019shaping}.

Single-agent POMDPs can be extended to the multi-agent setting by considering \emph{partially observable Markov games} (POMG)~\cite{Littman:1994:MGF:3091574.3091594, Hansen:2004:DPP:1597148.1597262}, a general and challenging category of learning problems for multi-agent systems. Very closely related notions can be found in the literature, for instance iPOMDPs \cite{gmytrasiewicz2005framework, han2018learning, doshi2009monte, han2019ipomdp}, and imperfect information games in game theory, see e.g.~\cite{brown2018depth}. In this setting, agents may have competing goals, the full state of the world is not observable, and the environment appears non-stationary to any individual, by virtue of the learning dynamics of their co-players. At each time-step $t$, players $i=1,\dots,n$ share the environment state $S_t$. They each receive an observation $Y_t(i)$ drawn from $p(Y_t(i)|S_t, i)$ and form a history $H_t(i) = (Y_{\leq t}(i), A_{\leq t-1}(i))$ used to compute an action $A_t(i)$ from their own policy $\pi_i(A_t(i)|H_t(i))$. They receive a reward $R_t(S_t, \mathbf{A}_t)$, where $\mathbf{A}_t$ denotes the joint action $(A_t(1),\ldots, A_t(n))$, and the common state changes according to $p(S_{t+1}|S_t, \mathbf{A}_t)$. 

Following the reasoning from POMDPs, it intuitively makes sense for each agent to compute a belief state representing their own uncertainty about the state of the environment. For reasons which will become clear shortly, we call such a belief state a \textit{zeroth-order belief}, and denote it $B^0_t(i)$ for agent $i$. From the perspective of a single agent, when the policies of other agents are fixed, the POMG appears as a POMDP: the other agents become part of the environment, and the unknown state of the world is therefore not only the environment state $S_t$, but also the state of knowledge of other agents. For that reason, agent $i$ has to form a belief about other agents' beliefs, which are \emph{first-order beliefs}. This leads to a recursion, where agents can now form \emph{second-order beliefs} about other agents' first-order beliefs, and so on. We define the belief of order $l$ for agent $i$ as the distribution of other agents' beliefs at order $l-1$ given information available to agent $i$:
\begin{align*}
    B^0_t(i) &\defeq p(S_t|H_t(i)) \, , \\
    B^{l}_t(i) &\defeq p\left(B_t^{l-1}(j)_{j \not = i}|H_t(i)\right) \text{ for } l\geq 1 \, .
\end{align*}
The question of how to represent and compute these belief states is even more prominent than in single-player games, given their recursive nature. As a tractable solution to this problem, we combine the idea of learning higher order beliefs in multi-agent games with that of representing belief states as neural codes. We detail the proposed method in the next section.

\subsection{Neural recursive belief states}\label{sec:rnbs}

We propose learning recursive belief models from agent experience. Following the principle of decentralized training (i.e.\ without a centralized critic) and decentralized execution, we assume that we have access to the environment state and agent histories in hindsight.\footnote{This is a common human learning behaviour. For instance, in a game of cards we could look back into played games and the hands that each player had.} At training time, we provide the agents with a handcrafted state, which captures important information about the full state (this is merely done to aid speed of learning). We re-define $S_t$ to denote the hand-crafted state. During training, we generate \textit{trajectories} from the $n$ players $\tau_t = (S_t, H_t(1), \ldots, H_t(n))$.

\textbf{Learning zeroth-order beliefs}. Agent $i$ learns $B^0_t$ by combining an RNN that computes the neural code $\hat{b}^0_t(i) \defeq f^b_\theta(\hat{b}^0_{t-1}(i), Y_t(i))$), with a parametric generative model that maps the neural belief state $\hat{b}^0_t(i)$ to a distribution $p_\phi(S_t|\hat{b}^0_t(i))$. The generative model can take any form and outputs a distribution in a very large space. The only requirement is efficient computation of its log likelihood $\log p_\phi(S_t|\hat{b}^0_t(i))$. This model can be learned by minimizing the following loss over $\theta$ and $\phi$,
\begin{align}
    \mathcal{L}^0(\theta, \phi) &= - \expec_{p(\tau_t)} \log p_{\phi}(S_t|\hat{b}^0_t(i)) \, ,
\end{align}
where $\hat{b}^0_t(i)$ is the neural belief state as a function of the entire history $H_t(i)$ and of $\theta$, and the gradient is computed using backpropagation through time. 

\begin{figure}[htb]
\subcaptionbox{A second-order generative model and its nested representation for a sampling factor $K=2$.\label{fig:generative_model}}[0.48\textwidth]{
    \includegraphics[width=0.48\textwidth]{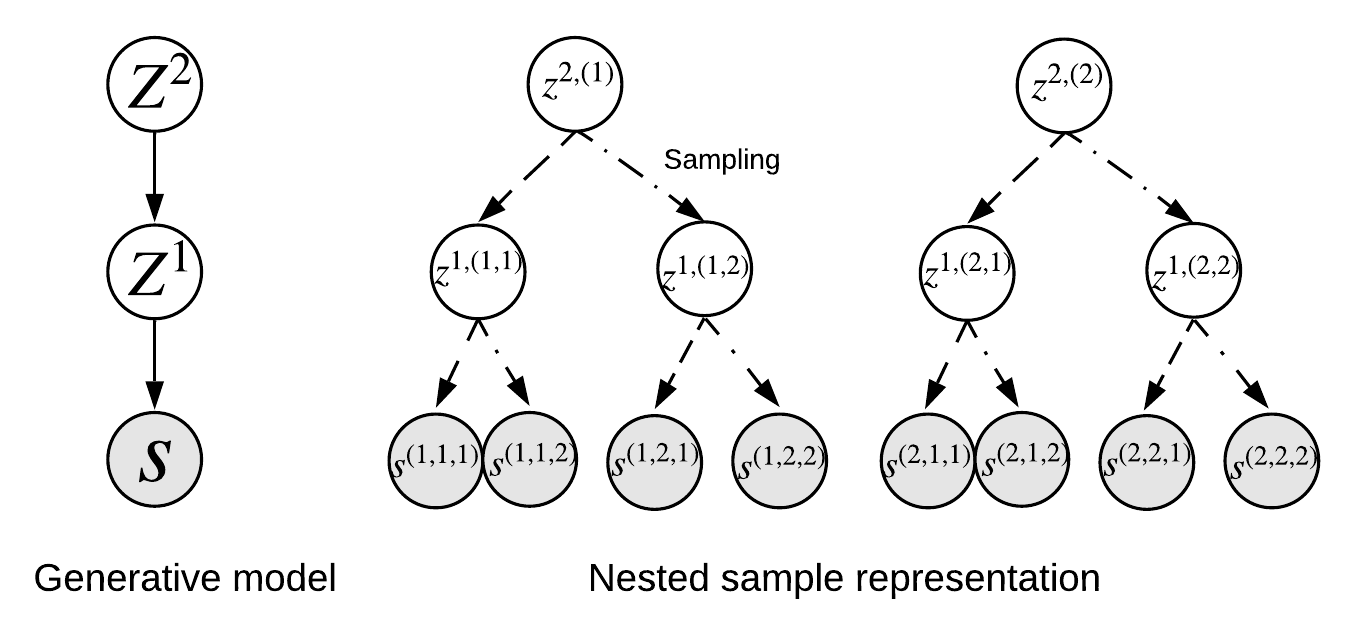}
    }
    \hspace{1em}
    \subcaptionbox{Agent with belief model. Black arrows are components of recurrent networks, red arrows denote belief models, blue arrows refer to policy and value. \label{fig:architecture} }[0.38\textwidth]{
    \includegraphics[width=0.38\textwidth]{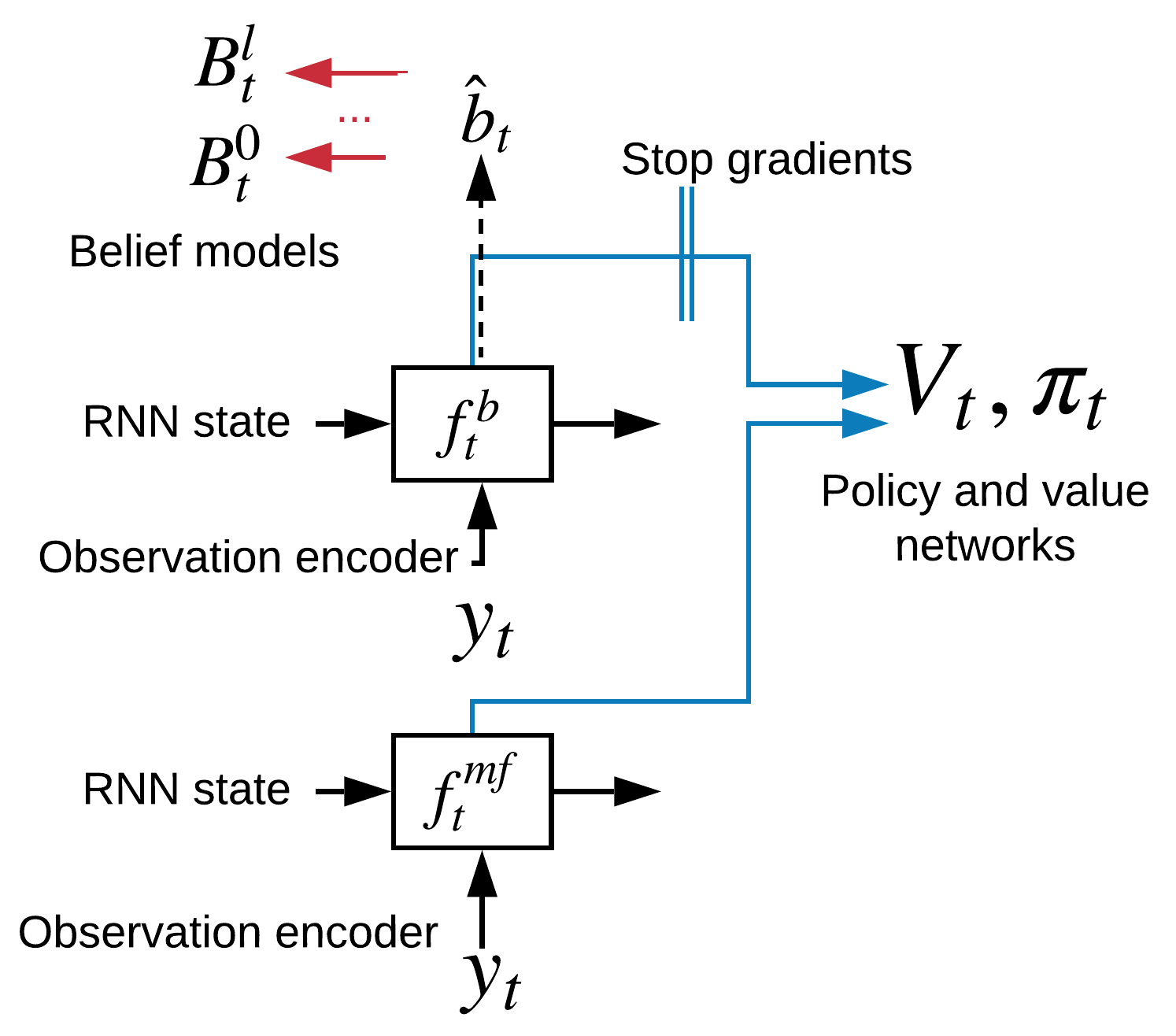}
 }
 \caption{Generative model and agent architecture.}
\end{figure}

\noindent \textbf{Learning higher-order beliefs}. In relation to theory of mind, our proposed approach is based on the following questions: (i) \emph{what would I have known if I had observed what the others have observed?}, followed by (ii) \emph{given what I have observed, what do I think the others know?} We first consider the case of two agents, and present the generalization to $n$ agents at the end of this section. For  simplicity, we drop the time-step index below. Consider a pair of agents $i$ and $j$. We must derive learning algorithm for the belief of agent $i$ of order $l$ (about the belief of agent $j$ of order $l-1$). Na\"ively, one might try to model the entirety of the knowledge agent $j$ may have, its entire history $H(j)$. While conceptually desirable, modeling entire histories of observations is computationally difficult. Moreover, it does not truly represent the state of knowledge inferred by agent $j$. Another option is to model the belief of order $l-1$ of agent $j$ directly, by learning a generative model $p(\hat{b}^{l-1}(j)|\hat{b}^l(i))$.  However, it is difficult to define a likelihood on the space of neural codes, as distances in that space are not indicative of the distance between the corresponding distributions. Instead we assume that the generative models corresponding to neural belief states at different orders can readily be sampled. Recall that a distribution can be empirically approximated by a collection of samples. A belief of order $0$ can be approximated by a collection of states $b^{0}(j) \defeq \left(s^{(1)}(j), \ldots, s^{(K)}(j)\right)$. A belief of order $1$ can in turn be approximated by a collection of collections of states. Following the recursion, beliefs of any order are represented by a nested collection, where the innermost constituents are collections of states. Agent $i$'s belief model of order $l$ is approximated by a collection of beliefs of order $l-1$ as $b^{l}(i) \defeq \left(b^{l-1,(1)}(j\ne i), \ldots, b^{l-1,(K)}(j \ne i)\right)$. Representing hierarchical distributions this way is a standard procedure in hierarchical Bayesian learning. When subsets of data are assumed to come from latent components, these are referred to as grouped data \cite{doi:10.1198/016214506000000302, Vandermeulen2015OnTI}. We propose learning hierarchical latent variable models, as these naturally capture the notion of distributions (of distributions, etc.). The latent variables can readily model the i.i.d.\ structure of nested samples that represent a target belief. Specifically, we employ a hierarchy of $l$ latent variables $Z_1, \ldots, Z_l$, defining 
\begin{equation}
    p_\phi(S|\hat{b}^l(i)) \defeq \int_{Z_1,\ldots, Z_l} p_\phi(Z^l | \hat{b}^l(i)) p_\phi(S|Z_1, \hat b^l(i)) \prod_{m=2}^l p_\phi(Z^{m-1}|Z^{m}, \hat b^l(i))
\end{equation}
as a model for the belief of order $l$. In order to generate a nested sample representation from such model, we follow the process illustrated in Fig. \ref{fig:generative_model} (note that the conditioning of the neural code is not shown). We first sample $K$ possible values for the variable $Z_l$, and, conditioned on each of the sampled values $z^{l, (k)}$, we again sample $K$ variables $z^{l-1}$. We recurse this process until all that remains is to sample $K$ state values from $p(S|z^1, \hat{b}^l(i))$.\footnote{Using the same sampling factor $K$ at every level is a particular choice made for simplicity, but one can use different values as needed. } It remains to explain how to learn the neural code of order $l$, which goes back to the two questions posed. An agent learns its belief models by having access to the history of the other agent. That is, the agent infers the neural code of agent $j$ by using its own internal RNNs on $H_t(j)$, and generates the nested sample $b^{l-1}(j)$. This refers to question (i). The learning objective is therefore to model $b^{l-1}(j)$ conditioned on its own neural code $\hat b^l(i)$, addressing question (ii). The learning objective relies on trajectories generated with samples from the belief models $\tau_t = (S_t, \tau_t(1), \ldots, \tau_t(N))$, where $\tau_t(i)=(H_t(i), b^1_t(i), \ldots, b^l_t(i))$. Formally, loss is then given by
\begin{align}
\mathcal{L}^{l}(\theta, \phi) &= - \expec_{p(\tau_t)} \log p_\phi(b^{l-1}(j)|\hat b^{l}(i)) \, .
\label{eq:order_l_loss}
\end{align}
This loss involves a marginalization of the latent variables which is intractable for all but the simplest cases. We can instead use variational inference by maximizing the evidence lower bound (ELBO), and use stochastic variational inference and the re-parameterization trick to obtain efficient gradients over $\phi$. This is known as the variational auto-encoder (VAE) \cite{conf/icml/RezendeMW14, kingmaW13}. The idea of using VAEs to learn distributions of distributions this way is related to the neural statistician \cite{DBLP:conf/iclr/EdwardsS17}. We describe further details regarding the objective function in Appendix \ref{sec:appendix_vae_beliefs}.
To summarize, $\hat b_t$ is an agent's \emph{neural recursive belief state}, an internal representation of the belief in the form of neural codes $\hat b_t \defeq (\hat b^{0}_t, \dots, \hat b^{l}_t)$. The code at order $l$ conditions the belief distribution $B_t^{l}$ over another agent's corresponding nested samples $b_t^{l-1}$. In the $n$-player case, it is the same method but involves an expansion of $n-1$ beliefs (in relation to each of the other agents) at every level of the recursion. Further discussion is in Appendix \ref{sec:appendix_n_players}. Note how the computational complexity scales exponentially as $\mathcal{O}\left((nK)^l\right)$. Fortunately, we can control the sampling factor according to computational availability, and in practice we expect strong diminishing returns from higher order reasoning. 

There are a number of benefits of our proposed sample-based approach. First, the belief models are ultimately grounded on a well-defined space (the state). Second, the method allows for an efficient decentralized implementation in which agents only require access to the histories of other agents (instead of requiring access to the  models themselves). Finally, it side-steps the intractability of solutions that rely on explicit belief filtering.

\noindent \textbf{The importance of multiple grouped samples}. A na\"ive version of our approach is to model nested single samples ($K=1$), instead of joint nested collections of samples. This method, however, can fail to capture the true distribution of the target beliefs, potentially limiting the ability of agents to compute optimal policies. To see how this applies to learning a first-order belief, consider the following version of the game of Tiger originally proposed as a toy example for single-agent POMDPs \cite{kaelblingLC98}. A player (P1) is faced with two rooms, one with a tiger and one with a prize.  The state of the game is the binary variable $S_t \in \{TL, TR\}$ (whether the tiger is behind the left or right door). P1 can open one door or wait and listen. When P1 chooses to listen, there is a $50\%$ chance of the tiger growling which reveals its door to P1. Clearly, P1 should listen until the tiger growls, and proceed to open the other door. On each round, a second player (P2) passively listens from further away, and can hear whether the tiger growled or not, but cannot tell from which door the growl came. Intuitively, when the tiger growls, P2 becomes certain that P1 is certain of where the tiger is; if instead P2 does not hear a growl, it knows with certainty that P1 is uncertain of the location of the tiger. Formally, P2 has a first-order model $B_t^1(\text{P2})$ of P1's model $B^0_t(\text{P1}) = p(S_t|H_t(\text{P1}))$. Crucially, if P2 learns $B_t^1(\text{P2})$ from one sample $s^{(1)}_t(\text{P1}) \sim B^0_t(\text{P1})$, it will learn over many episodes that 
\begin{align*}
p(B^0_t(\text{P1})=\text{TL}|Y_t(\text{P2})=\text{growl}) =  p(B^0_t(\text{P1})=\text{TL}|Y_t(\text{P2})=\text{silence}) &= 0.5  \, .
\end{align*}
This is because the marginal probability of the tiger's location is always $50\%$ on either door. Thus, P2 cannot know whether P1 will open a door or listen again if it uses this model. On the other hand, if P2 learns to model a collection of samples ${(s_t^{(1)}(\text{P1}),\dots, s_t^{(K)}(\text{P1}))}$, then P2 will know that 
\begin{align*}
&p\left(B_t^{0,(1)}(\text{P1})=\cdots=B_t^{0,(K)}(\text{P1})=TL|Y_t(\textrm{P2})=\text{growl}\right)=0.5, \text{ and} \\ 
&p\left(B_t^{0,(1)}(\text{P1})=\cdots=B_t^{0,(K)}(\text{P1})=TL|Y_t(\text{P2})=\text{silence}\right)=0.5^K \, ,
\end{align*}
since hearing the growl introduces conditional dependence. Because the probabilities are different depending on whether the tiger growls or not, P2 forms the correct belief of P1's belief. This would allow P2 to predict whether P1 would chose to listen or not. In our experiments, we implement this game as an RL environment and train agents that showcase the need for learning correct beliefs in order to learn optimal policies. In Appendix \ref{sec:appendix_tiger_b2}, we also extend this game to a version with a third player that has to form a second order belief.

\subsection{Agent architectures}
While above we defined a different neural code for every belief order, in practice $\hat b_t$ can be implemented by a single code. The agent architecture used in the experiments is an extension to a standard deep architecture with an RNN: it is composed of two recurrent components which aggregate the history of observations to produce a representation as shown in Fig.\ \ref{fig:architecture}. By separating the internal representation into two, a model-free part and a belief representation part, we decouple interactions between the gradient signals of the RL and belief losses which we empirically saw to be helpful. \footnote{The ``stop gradient'' components ensure this decoupling during the learning phase. Both recurrent states are concatenated to produce the complete agent representation for the policy and value prediction.} The two \textit{observation encoders} are composed each of a convolutional neural network (CNN), connected to a multi-layered perceptron (MLP). The encoded observations are inputs to the RNN, which are implemented as GRUs \cite{ChoMGBBSB14}. The outputs of each RNN serve as inputs to both the belief models as well as the policy and value components. The latter are MLPs that produce the action probabilities and the critic of the current state. Specific details of the networks architecture is detailed in Appendix \ref{sec:appendix_agent_architectures}.

\section{Experiments}\label{sec:experiments}

\subsection{Environments}

Our aim is to evaluate the usefulness of using belief models in MARL agents. To do so we describe two environments below, and provide further details and illustrations in Appendix \ref{sec:appendix_games}.

\textit{Running with Scissors} is a partially observable, pure-conflicting interest task extending the classic matrix game rock-paper-scissors to an intertemporal gridworld environment \cite{vezhnevets2019options}. Two players move around a small environment ($15\times23$) and collect resources. Resources have three types (rock, paper, and scissors), which are visually distinguishable from one another. To maximize the reward, players should correctly identify which resource the opponent is collecting and then collect as many counter-resources as possible. Each episode automatically times out after 1000 steps. The hand-crafted state used to build the belief models is defined to be the (private) collected number of rocks, papers, and scissors.

We formalize the \textit{Tiger game} described above as an RL environment. The first player P1 has to choose one of two doors: one has a prize (reward of 1), and the other one a tiger (-5 reward). At every step, the player has 3 actions, open left door (OL), open right door (OR) or listen (L). Listening gives no reward, and the game ends when a door is opened. The task of P2 is to guess at each round whether P1's action is to open a door or to wait and listen based on its observation (with a reward of 1 with the correct guess, or 0 otherwise).

\subsection{Training setup}
The training regime in Running with Scissors is based on \emph{multiple population play}, where a number of agents are sampled from different sub-populations to play in each game. Each sub-population consists of a number of agents of a similar type, each of which has an independent set of network parameters. For each episode, the required number of agents for the game are randomly chosen from the pool of all agents. We distinguish three sub-populations. The \textit{specialist agents} (S), which are trained with a combination of the task reward and pseudorewards designed to produce interpretable policies. In Running with Scissors, these include an agent with high preference for collecting rocks, one for collecting scissors, and one for paper. These specialist agents receive a pseudoreward of $+5$ for collecting the item of their preference. The second sub-population of agents are \emph{non-specialist} agents (NS), which simply learn to maximize the task rewards. The third sub-population comprises the \textit{main agents} we wish to evaluate. The type will vary in every experiment so we can compare the types under the same training regime. These types include agents that use the $B^0$ belief model (B0 agents), agents with first order belief models (B1 agents), and two types as baselines. First is a sub-population of model-free agents in which the architecture of the agent is the same as the agents with belief models, but both RNNs are used solely to estimate the policy and value (these agents also have belief models, but the gradients from belief models do not update the RNN parameters). This allows us to have a fair comparison of model-free and belief agents in terms of number of parameters (with the model-free baseline having overall more trainable parameters for the policy due to the second RNN). The second baseline involves agents that has a predictive model of the opponents trajectory (Tr.\ agent). These agents learn to predict the actions and visual observations of the other agents at every time-step using the second RNN (details in Appendix \ref{sec:appendix_trajectory_agent}). The difference in performance between B0 and Tr.\ agents can highlight the advantage of predicting the salient hand-crafted states in contrast with predicting the opponents' actions and observations.

The training process for both environments uses the distributed actor-critic setup of the Importance Weighted Actor-Learner Architecture \cite{impala2018} (but other RL algorithms are also compatible). Each actor generates trajectories, which are sent to the learner in chunks of 100 environment steps. The learner then optimizes the RL and belief losses jointly. Further details on hyper-parameters used are given in Appendix \ref{sec:appendix_agent_architectures}.

\begin{figure}[ht]
     \centering
     \begin{subfigure}[t]{0.23\textwidth}
         \centering
         \includegraphics[width=\textwidth]{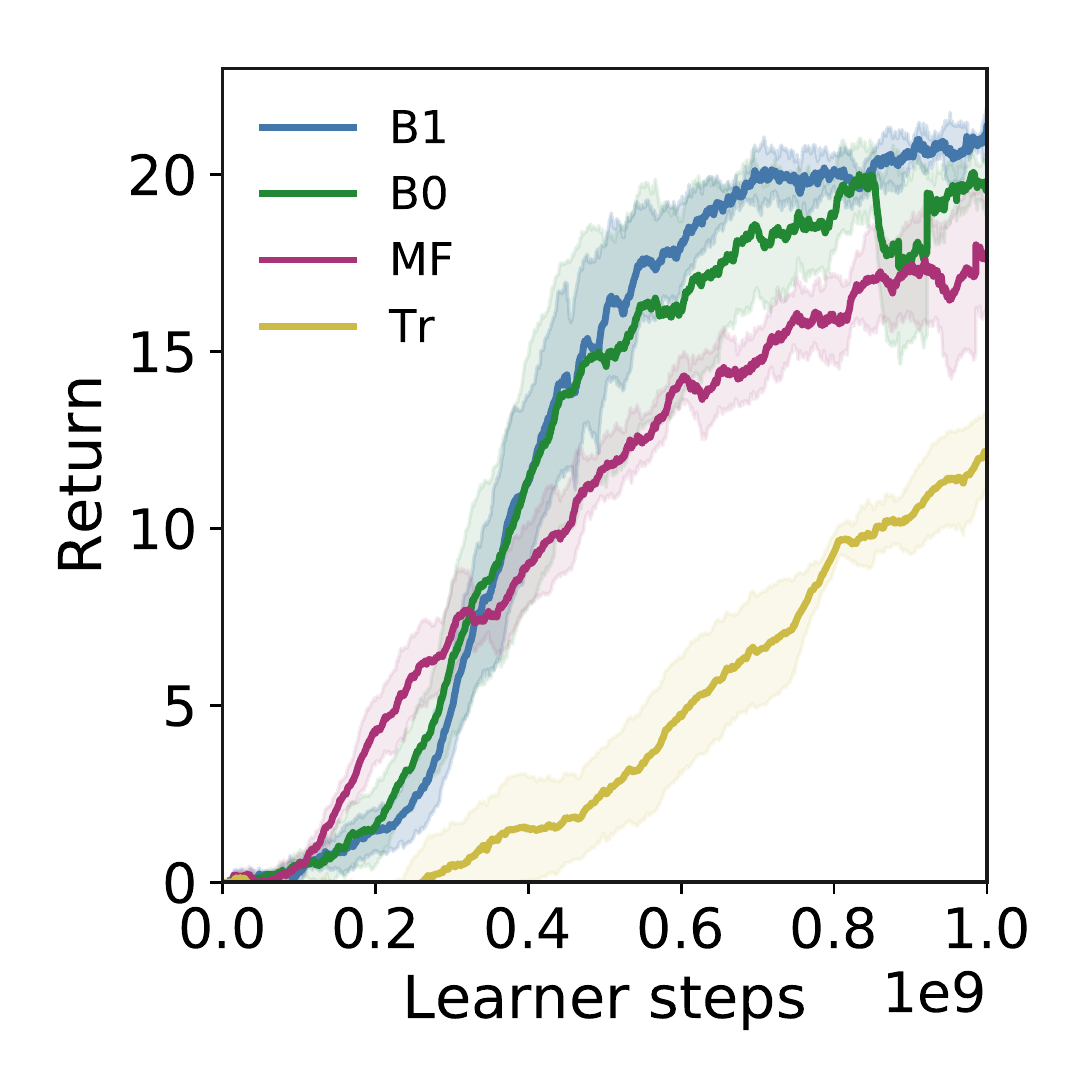}
         \caption{Main agent performance in experiments with 3 S, 3 NS and 1 main agent, per main agent type (6 independent runs). }
         \label{fig:rws_7joint}
     \end{subfigure}
     \hspace{0.5em}
     \begin{subfigure}[t]{0.23\textwidth}
         \centering
         \includegraphics[width=\textwidth]{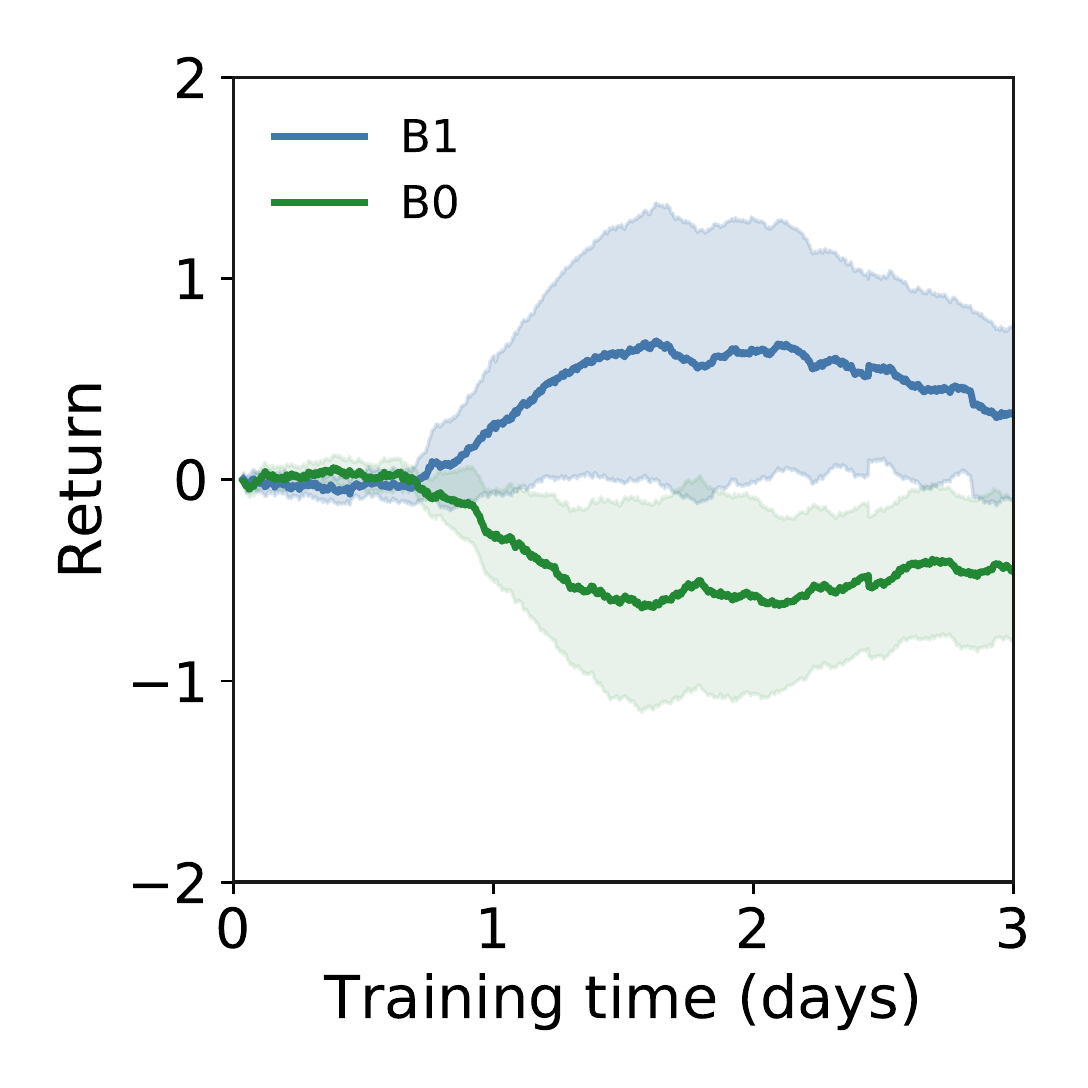}
         \caption{Average return of experiments with 3 B0 and 3 B1 agent populations (8 independent runs).}
         \label{fig:rws_b0vb1}
     \end{subfigure}
     \hspace{0.5em}
     \begin{subfigure}[t]{0.23\textwidth}
         \centering
         \includegraphics[width=\textwidth]{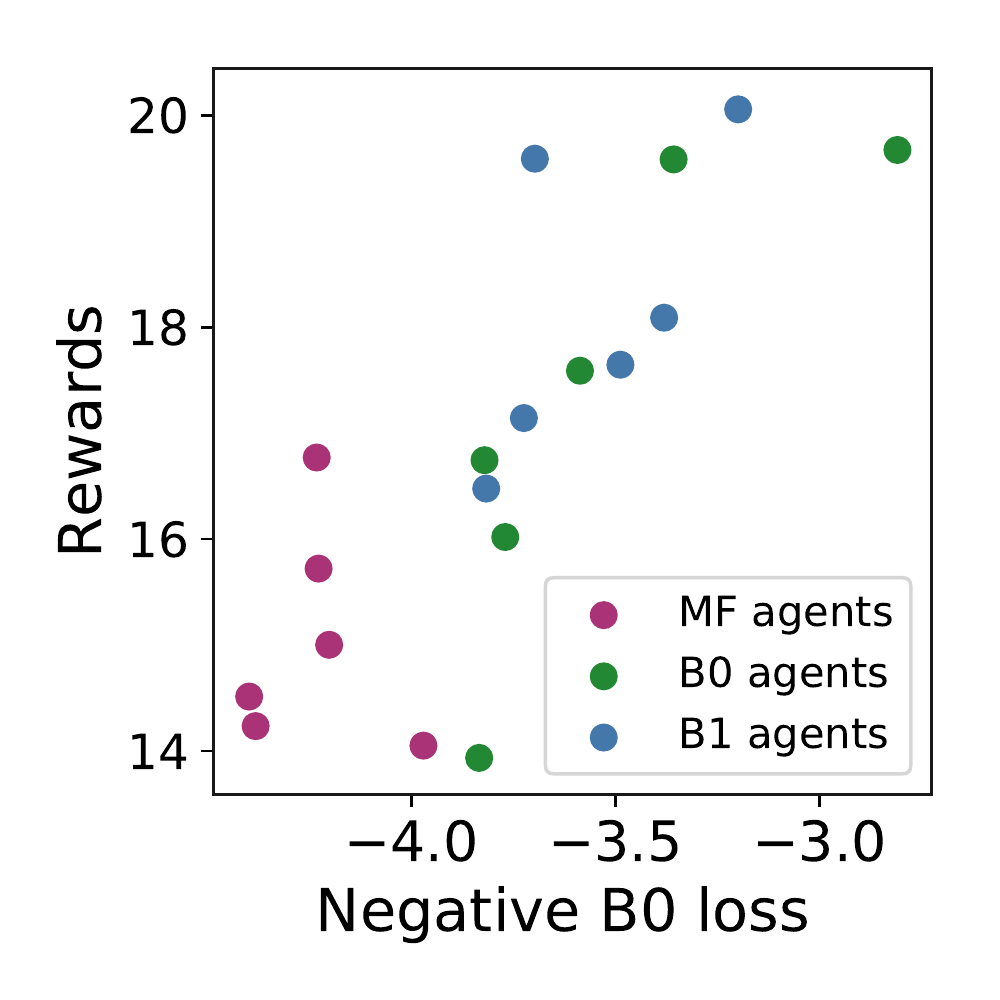}
         \caption{$L^0$ loss and average rewards correlations on the experiments with 3 S, 3 NS and 1 main agent, per main agent type.}
         \label{fig:rws_symmetric_scatterb0}
     \end{subfigure}
     \hspace{0.5em}
     \begin{subfigure}[t]{0.23\textwidth}
         \centering
         \includegraphics[width=\textwidth]{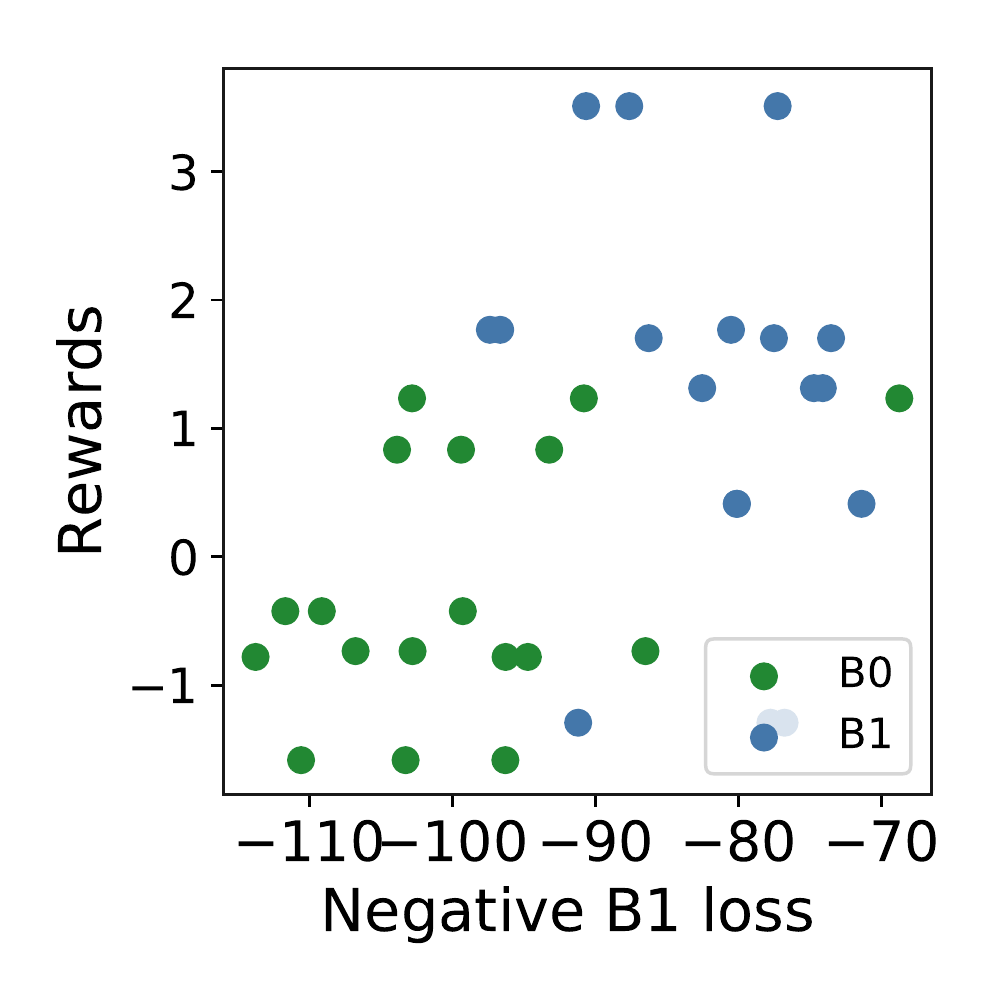}
         \caption{$L^1$ loss and average rewards correlations on the experiments with 3 B0 and 3 B1 agents populations.}
         \label{fig:rws_symmetric_scatterb1}
     \end{subfigure}  
    \caption{Training results in Running with Scissors. Shaded areas represent the 95\% confidence intervals for the means averaged over a number of experiment runs.}
    \label{fig:rws_results}
\end{figure}

\noindent \textbf{Running with Scissors}.  We first experiment with a multiple population set-up which has three specialist agents, three non-specialist, and one main agent for the third sub-population.  Fig.\ \ref{fig:rws_7joint} shows the average return obtained by the different types of of the main sub-population. We see how both agents with $B^0$ and $B^1$ beliefs clearly outperform the baselines. Also, predicting the opponent's actions and observations does not provide useful representations for the task as seen by the poor performance of the agent with a trajectory-prediction model. To test the hypothesis that higher order reasoning should outperform lower order, we run an experiment which involves two populations of three B0 agents and three B1 agents. We observe in Fig.\ \ref{fig:rws_b0vb1} how B1 agents achieve higher average return than the lower-order counterpart. We also analyze the correlation between average return obtained by these different types and their average belief losses. As we would expect, better belief predictive models lead to higher rewards. Fig.\ \ref{fig:rws_symmetric_scatterb0} shows the correlation with respect to 0th order belief loss. Fig.\ \ref{fig:rws_symmetric_scatterb1} shows the 1st order belief loss for the B0 vs. B1 agents. These results suggest that both belief models enable the agents to infer useful information from their experience.
In summary, we see agents with belief representations outperform the baselines in Running with Scissors by identifying the strategies of their opponent in real time and deploying counter-strategies. 

\noindent \textbf{Tiger game}. The Tiger game proposed above is a toy problem designed to  conceptually highlight the potential issues of incorrect modelling of beliefs. We  therefore empirically evaluate whether our representation learning method enables solving the games by computing the correct beliefs.

\begin{figure*}[!htb]
     \centering
     \begin{subfigure}[b]{0.33\textwidth}
         \centering
         \includegraphics[width=\textwidth]{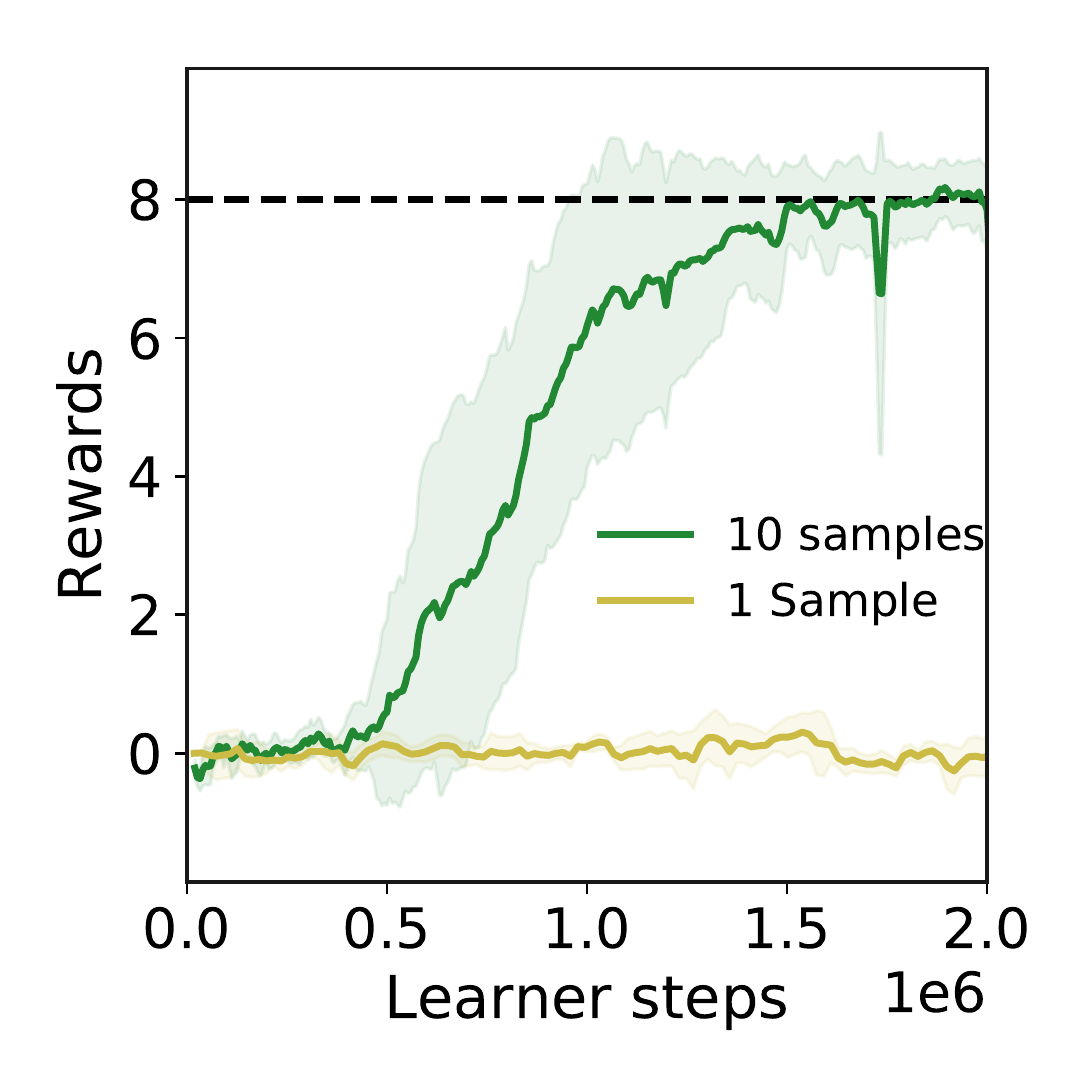}
         \caption{Average return obtained by P2 agents trained with either a belief model learned from K=10 or K=1 samples (from the B0 model of P1).}
         \label{fig:tiger_b1}
     \end{subfigure}
     \hspace{0.5em}
     \begin{subfigure}[b]{0.33\textwidth}
         \centering
         \includegraphics[width=\textwidth]{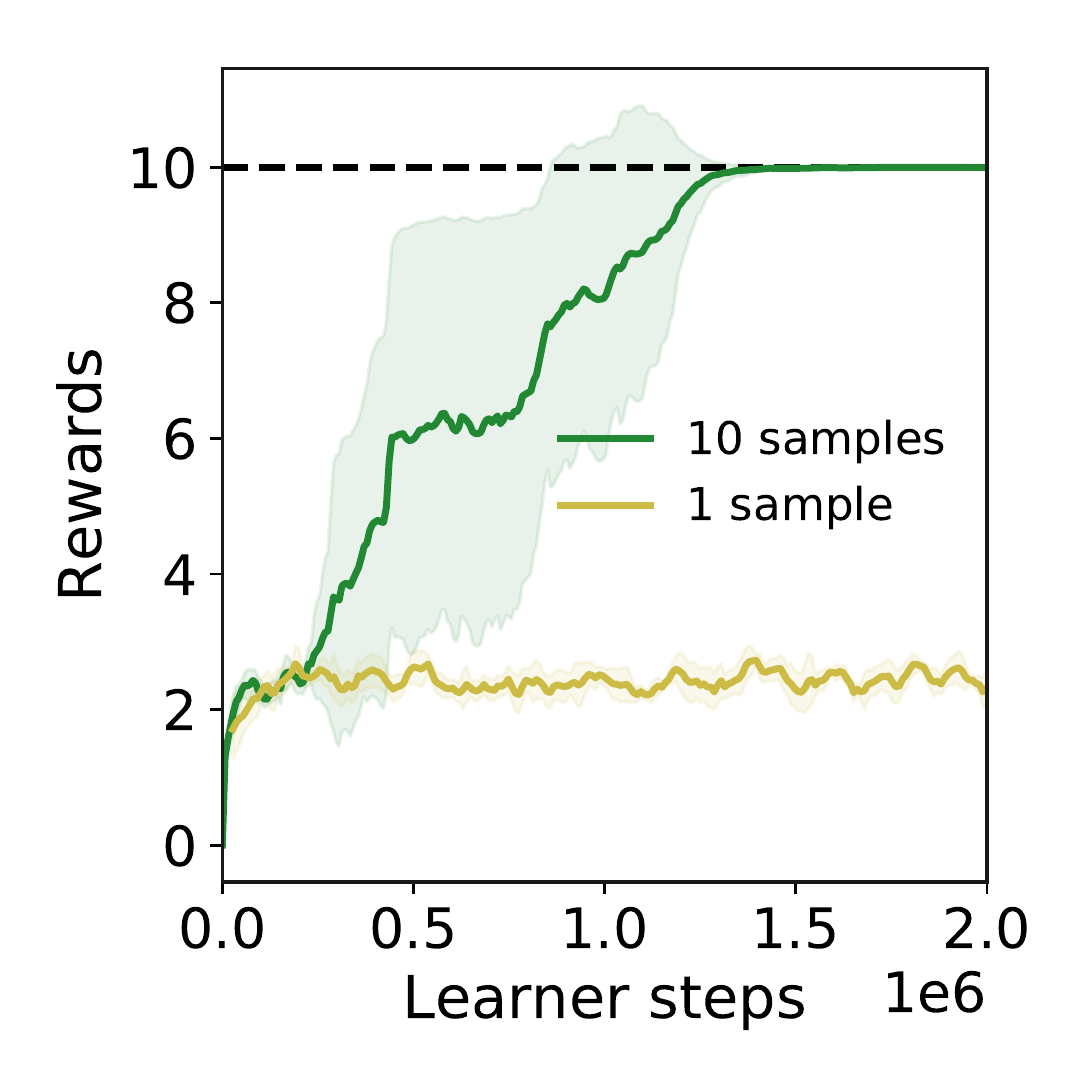}
         \caption{Average return obtained by P3 agents trained with either a belief model learned from K=10 or K=1 samples (of the B1 model of P2).}
         \label{fig:tiger_b2}
     \end{subfigure}
        \caption{Tiger experimental results when playing against optimal players and optimal beliefs. The optimal policy is denoted by a black dashed line.}
        \label{fig:tiger_results}
\end{figure*}

We experiment with the two versions of the Tiger game to evaluate the ability to learn agents' beliefs and use these beliefs to learn the policies via RL. We run these experiments following the same training method and architecture as detailed in Appendix \ref{sec:appendix_agent_architectures}. 
In the first experiment, we match P2 with an ``optimal'' P1 in the version of the game with two players. The input to the policy and value functions for P2 is thus composed of samples $b_t^1(P2)$ from its own belief model (thus, if the model is correct, the policy should be able to learn how to map these samples to the right actions).
In the second experiment, we evaluate the second version of Tiger with three players. We match P3 against an optimal P2, and learn the policy and value conditioned on $b_t^2(P3)$. We observe in Fig.\ \ref{fig:tiger_results} how in both cases, only the agents that learn from $K=10$ samples can solve the problem, while agents that learn from $K=1$ samples (i.e.\ theoretically unable to learn the correct higher order beliefs) catastrophically fail.

\begin{figure}[!htb]
     \centering
     \begin{subfigure}[t]{0.33\textwidth}
         \centering
         \includegraphics[width=\textwidth]{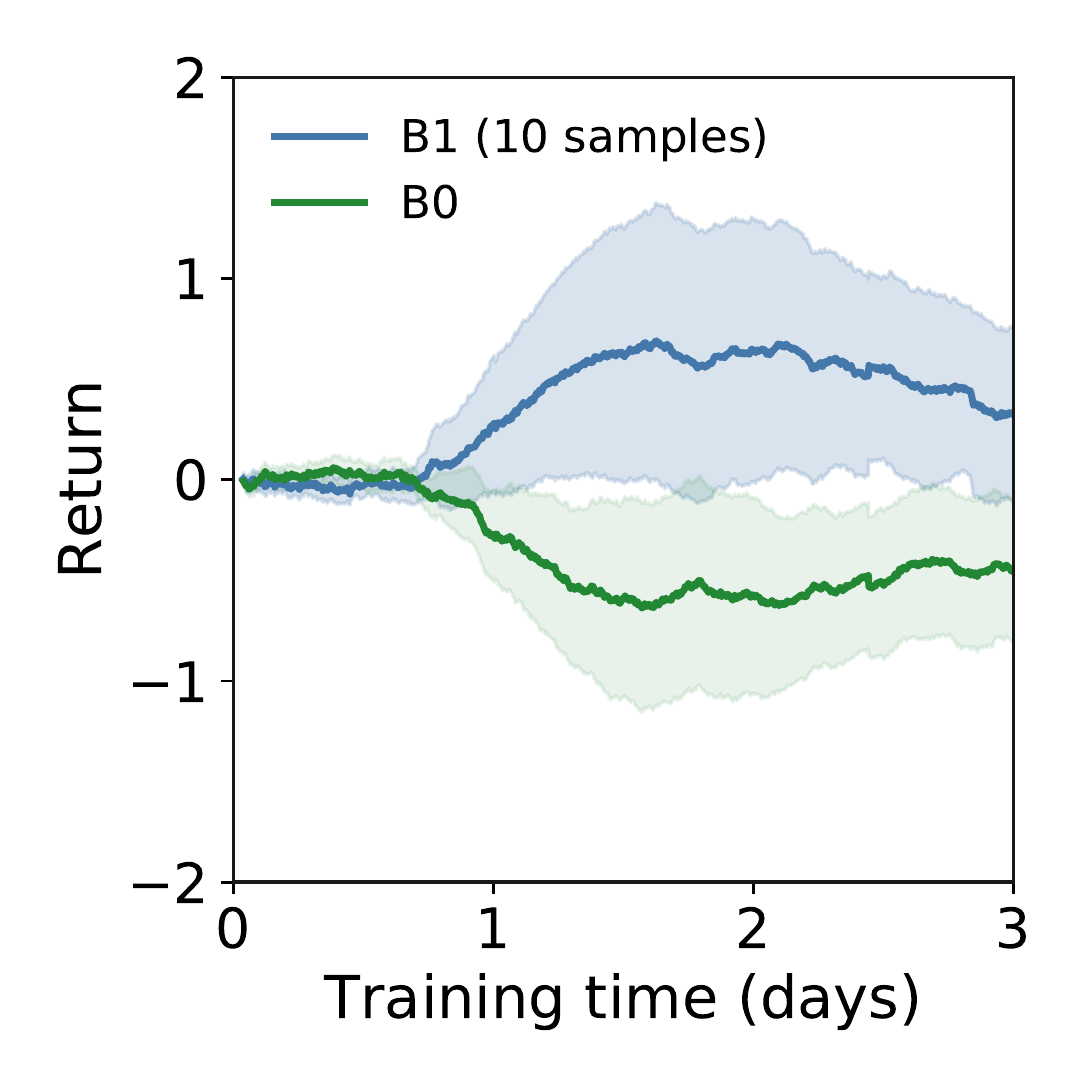}
         \caption{Average return of experiments with 3 B0 and 3 B1 ($K=10$) agent populations (8 independent runs).}
         \label{fig:rws_b0vb1_repeat}
     \end{subfigure}
     \hspace{0.5em}
     \begin{subfigure}[t]{0.33\textwidth}
         \centering
         \includegraphics[width=\textwidth]{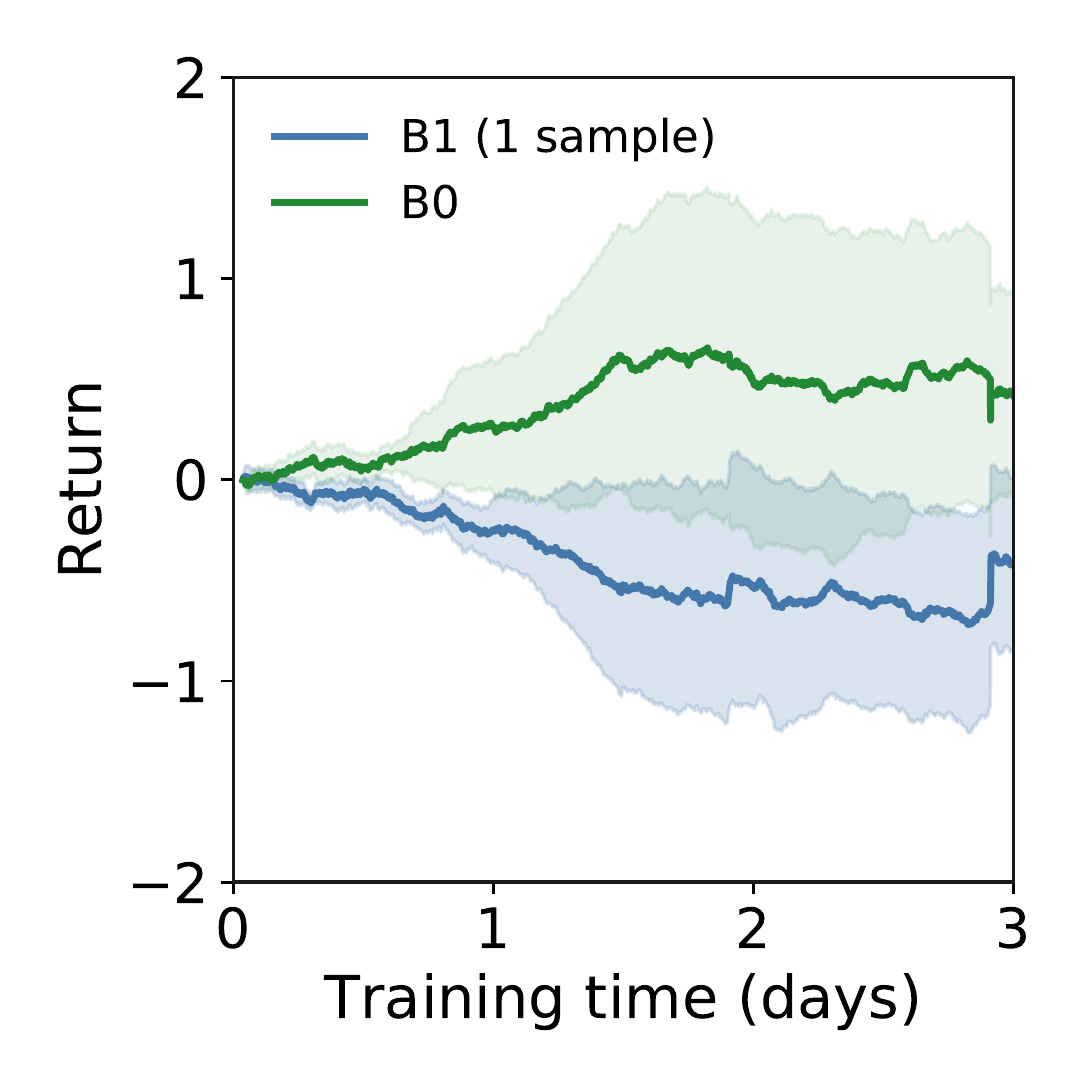}
         \caption{Average return of experiments with 3 B0 and 3 B1 ($K=1$) agent populations (8 independent runs).}
         \label{fig:rws_b0vb1_1sample}
     \end{subfigure} 
        \caption{Training results in Running with Scissors. Shaded areas represent the 95\% confidence intervals for the means averaged over a number of experiment runs.}
        \label{fig:b0_b1_samples}
\end{figure}

\noindent \textbf{Effect of the nested sampling factor $K$.} Above we showed that in the Tiger game, it becomes impossible to learn the correct distribution with one grouped sample at a time. While conceptually we know that the larger the sampling factor, the better the distributions are approximated, we also investigate in Running with Scissors the empirical effect of representing the nested beliefs with $K=1$, in contrast to a larger value $K=10$. We run experiments in Running with Scissors, where we compare the performance of an agent's first order reasoning using $K=10$, with one using $K=1$. We observe in Fig.\ \ref{fig:rws_b0vb1_repeat} how the B1 agents achieve higher average returns than the B0 agent. However, Fig.\ \ref{fig:rws_b0vb1_1sample} shows that if B1 models are only learned with one sample at at time, B1 agents fail to outmatch the B0 agents, and in fact seem to be doing slightly worse (with a larger variability in performance across 8 runs).

\section{Related work}\label{sec:rel-work}

The state-of-the-art in learning and leveraging belief models in multi-agent deep RL is the PR2 agent \cite{DBLP:conf/iclr/WenYLWP19}. This algorithm learns a model of others' policies and incorporates this into its update step based on Bayesian reasoning, which leads to the discovery of optimal equilibria in simple coordination problems like matrix games and particle worlds. However, the method is computationally expensive, and it is not clear how well it would scale to more complex tasks with populations of co-players and competitive dynamics. Similarly the BAD \cite{DBLP:conf/icml/FoersterSHBDWBB19} and SAD \cite{anonymous2020simplified} algorithms learn belief models, but are specialized to fully cooperative games. Theory of mind models were proposed in the context of agents with social influence \cite{DBLP:journals/corr/abs-1810-08647}, leading to emergent communication and cooperation in social dilemmas. Unlike our model, successfully leveraging these beliefs requires an intrinsic motivation. The work of \cite{vezhnevets2019options} proposes a hierarchical agent (OPRE), applied also to Running with Scissors, that discovers options useful for strong generalization. 
Our work does not address generalization but rather focuses on
learning representations that support higher opponent awareness and, as a consequence, better decision making.
Our methods are thus not directly comparable; in fact, the latent options discovered by OPRE could be used alongside the representations learned to capture higher order reasoning. 

In single-agent RL, the two primary research aims of previous work in learning belief models have been representation richness and improved task performance in POMDPs. This applies both to works that investigate unsupervised learning techniques \cite{venkatraman2017predictive,IglZLWW18,buesing2018learning,guo2018neural,gregor2019shaping} or leverage supervision on state information  \cite{moreno2018neural,humplik2019meta}. These works, however, do not address the multi-agent setting. 

The field of epistemic game theory studies the reasoning of agents and their beliefs in a formal manner which has lead to a deeper understanding of these notions and to novel solution concepts, see e.g.\ \cite{RePEc:wsi:igtrxx:v:16:y:2014:i:01:n:s0219198914400015} for a historical account. However, the analyses in this field do not usually provide practical solutions to RL in complex environments. Our work is related to the interactive POMDP (iPOMDP) formalism \cite{gmytrasiewicz2005framework} with respect to the notion of recursive belief states, but the intractable formulation of a iPOMDP limit its application to small-scale problems. While some works propose approximations to improve the applicability of iPOMDPs \cite{doshi2009monte, han2019ipomdp}, these methods still do not address the intractability of having to explicitly compute the belief updates.

\section{Conclusion}
We have investigated the use of deep generative models to learn the recursive belief structures that emerge in POMGs. The proposed approach offers a practical framework for learning belief models that are useful in complex environments as shown in our experiments. In future work, we wish to evaluate in greater detail the empirical effects of the sample-size in learning the belief models, and to understand the effects of incorporating higher-order belief models ($B^2$ and higher), which may be important in scenarios requiring deep theory of mind, such as Hanabi \cite{DBLP:journals/corr/abs-1902-00506}. 
\bibliographystyle{plain}
\bibliography{biblio.bib}
\clearpage
\appendix

\section{Belief representation for n agents.}
\label{sec:appendix_n_players}

\begin{figure}[!ht]
    \centering
    \begin{tikzpicture}[minimum size=1.2cm]
  \tikzstyle{box}=[rectangle,rounded corners=4pt,minimum size=3em]
  
  

  \begin{scope}[every node/.style={box,draw,thick}, node distance=1.5cm]
      \node (b21) {$b^{2}(1)$};
      \node (b1) [below=of b21.base, anchor=base] {$b^{1}(2), \ldots, b^{1}(n)$};
      \draw[dashed] ($(b21.west)+(3.375ex,-1.5ex)+(-2ex,0ex)$) -- ($(b1.north west)+(1.5pt,-1.5pt)$); 
      \draw[dashed] ($(b21.west)+(3.375ex,-1.5ex)+(2ex,0ex)$) -- ($(b1.north east)+(-1.5pt,-1.5pt)$);
      
      \node (b11) [below=of b1.west, anchor=west] {$b^{1,(1)}(2), \ldots, b^{1,(K)}(2)$};
      \draw[dashed] ($(b1.west)+(3.375ex,-1.5ex)+(-2ex,0ex)$) -- ($(b11.north west)+(1.5pt,-1.5pt)$); 
      \draw[dashed] ($(b1.west)+(3.375ex,-1.5ex)+(2ex,0ex)$) -- ($(b11.north east)+(-1.5pt,-1.5pt)$);

      \node (b0) [below=of b11.west, anchor=west] {$b^{0}(1), b^{0}(3), \ldots, b^{0}(n)$};
      \draw[dashed] ($(b11.west)+(4.5ex,-1.5ex)+(-3.25ex,0ex)$) -- ($(b0.north west)+(1.5pt,-1.5pt)$); 
      \draw[dashed] ($(b11.west)+(4.5ex,-1.5ex)+(3.25ex,0ex)$) -- ($(b0.north east)+(-1.5pt,-1.5pt)$);
      
      \node (b01) [below=of b0.west, anchor=west] {$b^{0,(1)}(1), \ldots, b^{0,(K)}(1)$};
      \draw[dashed] ($(b0.west)+(3.375ex,-1.5ex)+(-2ex,0ex)$) -- ($(b01.north west)+(1.5pt,-1.5pt)$); 
      \draw[dashed] ($(b0.west)+(3.375ex,-1.5ex)+(2ex,0ex)$) -- ($(b01.north east)+(-1.5pt,-1.5pt)$);

      \node (s)  [below=of b01.west, anchor=west] {$s^{(1)}, \ldots, s^{(K)}$};
      \draw[dashed] ($(b01.west)+(4.5ex,-1.5ex)+(-3.25ex,0ex)$) -- ($(s.north west)+(1.5pt,-1.5pt)$); 
      \draw[dashed] ($(b01.west)+(4.5ex,-1.5ex)+(3.25ex,0ex)$) -- ($(s.north east)+(-1.5pt,-1.5pt)$);
  \end{scope}

\end{tikzpicture}
    \caption{Nested sample representation of the belief of order $2$ of agent 1, over an $n$-player game.}
    \label{fig:nested-sample}
\end{figure}
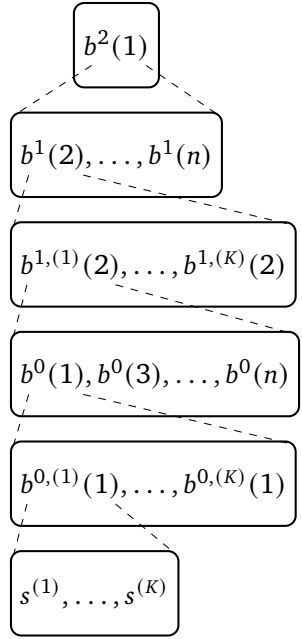

From the perspective of agent $i$, a POMG with $n$ agents can be modelled as in the two player case by substituting $j$ (as described in Sec.\ \ref{sec:belief_models}) by the set of $n-1$ agents omitting $i$. Assuming some arbitrary ordering of those agents, a belief of order $l$ is now a distribution of $n-1$ beliefs of order $l-1$. Fig.\ \ref{fig:nested-sample} illustrates the full nested representation with $n$ players and $K$ samples per order.

In terms of capturing this structure with a latent variable model, the generative model is now composed of $n-1$ latent variables at each order, where each represents the belief of a particular player at that order. That is, $p(Z^l|\hat b^l(i))$ becomes $p(Z^l(n\neq i)|\hat b^l(i))$, where $Z^l(n\neq i)$ indicates a sequence of $n-1$ latent variable vectors, and like-wise, $p(Z^{l-1}|Z^l(n), \hat b^l(i))$ becomes $p(Z^{l-1}(m\neq n))|Z^l(n), \hat b^l(i))$.
We reiterate that because of our theory of mind inspired approach, agent $i$ learns the belief models at every order using its own experience as well as the other agents' experience. This way, it can generate samples from a lower order belief model from the other agent's experience  in order to learn its higher order belief model, and so forth. However, at the highest designated belief order, $l$, the agent does not need to optimize the belief models using other agents' trajectories (as there is no need to generate samples anymore), and we can save the additional computation.

\section{Objective function for higher order beliefs.}
\label{sec:appendix_vae_beliefs}

The objective for a belief of order $l$, as per Eq. \ref{eq:order_l_loss}, is the minimization of the negative log probability of the belief samples (dropping the time index):

\begin{align*}
L^{l}(\theta, \phi) &= - \expec_{p(\tau)} \log p_\phi(b^{l-1}(j)|\hat b^{l}(i))
\end{align*}
 
As mentioned, we choose to use a variational objective as it has become a prominent and flexible method for learning latent variable models, and because it can handle the i.i.d.\ structure of nested samples in a simple and elegant way. Due to the hierarchical structure of the model, we can write the ELBO in a recursive way:
\begin{align}
\hat{L}^{l}(\theta, \phi) &= -\expec_{\tau}\ell_{E}(l, b^{l-1}(j), \hat b^l(i))
\label{eq:order_l_elbo_loss}
\\
\ell_{E}(l, b^{l}(j), Z^{l+1}, \hat b^{l'}(i)) &= \expec_{q_\phi(Z^{l}|b^{l}(j), \hat b^l(i))}\left[ \log \frac{q_\phi(Z^{l}|b^{l}(j), \hat b^{l}(i))}{p_\phi(Z^{l}|Z^{l+1}, \hat b^{l'}(i))} + \sum_k^K L_{E}(l-1, b^{l, (k)}(j), Z^{l, (k)}, \hat b^{l'}(i)) \right]
\label{eq:order_l_elbo} \\
\text{ for $l>1$, and} \nonumber\\
\ell_{E}(l=1, b^{1}(j), Z^{2}, \hat b^{l'}(i)) &= \expec_{q_\phi(y^{1}|b^{1}(j), \hat b^l(i))}\left[ \log \frac{q_\phi(Z^{1}|b^{1}(j), \hat b^{l'}(i))}{p_\phi(Z^{1}|Z^2, \hat b^{l'}(i))} +  \sum_k^K \log p_\phi(b^{1, (k)}(j)|Z_1, \hat b^{l'}(i))\right].
\label{eq:order_1_elbo}
\end{align}

\begin{figure}
    \centering
    \includegraphics[width=0.99\textwidth]{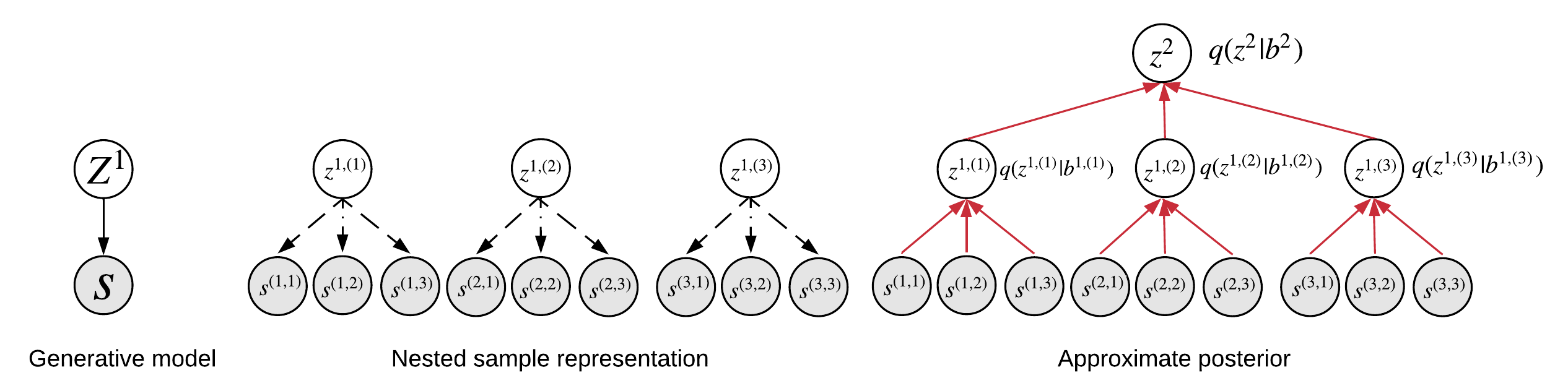}
    \caption{Example of a second order generative model and its nested representation for a sampling factor K=3, and the encoding of a nested representation to compute the approximate posterior.}
    \label{fig:nested_posterior}
\end{figure}

Note that $\theta$ and $\phi$ are implicitly conditioning the corresponding components (RNNs and parametric generative models). The approximate posterior is defined recursively in a bottom-up fashion from the inner-most nested subset of state variables, as illustrated in Fig.\ \ref{fig:nested_posterior} (right). For $l=2$, the full posterior would be $q(Z^{2}, Z^{1,(1:K)}|b^2) = q(Z^{2}|Z^{1,(1:K)})\prod_{k=1}^K q(Z^{1,(k)}|(b^{2,(k)})$, where each $Z^{1,(k)}$ is modelled using the same parametric distribution. An advantage of this variational formulation is that the approximate posterior can leverage the i.i.d.\ property of the samples by implementing each posterior distribution using an order invariant function of samples from the level below.

Above we described the general form of our higher order beliefs modelled as VAEs. While VAEs are generally very applicable to many settings, care needs to be taken so that these models learn well. Below we specify a number of techniques we incorporate, although we do not claim these to be necessarily the only or best methods available.

First, we notice how the ELBO objective is sensitive to the choice of $K$. In particular, the relative weight of the likelihood terms and the Kullback–Leibler (KL) divergence terms changes exponentially with $K$. This is especially relevant for higher order models due to the exponential growth in number of nested samples, and can lead for instance to the terms related to the higher level latent variables being ignored. To mitigate this issue, we choose to normalize the summation terms by $1/K$:

\begin{align}
\ell_{E}(l, b^{l}(j), Z^{l+1}, \hat b^{l'}(i)) &= \expec_{q_\phi(Z^{l}|b^{l}(j), \hat b^l(i))}\left[ \log \frac{q_\phi(Z^{l}|b^{l}(j), \hat b^{l'}(i))}{p_\phi(Z^{l}|Z^{l+1}, \hat b^{l'}(i))} + \frac{1}{K}\sum_k^K \ell_{E}(l-1, b^{l, (k)}(j), Z^{l, (k)}, \hat b^{l'}(i)) \right]
\label{eq:order_l_elbo_div_k} \\
\text{ for $l>1$, and} \nonumber\\
\ell_{E}(l=1, b^{1}(j), Z^{2}, \hat b^{l'}(i)) &= \expec_{q_\phi(y^{1}|b^{1}(j), \hat b^{l'}(i))}\left[ \log \frac{q_\phi(Z^{1}|b^{1}(j), \hat b^{l'}(i))}{p_\phi(Z^{1}|Z^2, \hat b^{l'}(i))} +  \frac{1}{K}\sum_k^K \log p_\phi(b^{1, (k)}(j)|Z_1, \hat b^{l'}(i))\right],
\label{eq:order_1_elbo_div_k}
\end{align}

We note that changing the weighting between the KL and likelihood terms is practiced for different purposes, see e.g.\ $\beta\text{-VAE}$ \cite{HigginsMPBGBML17} and generalized Variational Inference \cite{DBLP:journals/corr/abs-1904-02063} for a more detailed discussion.

Finally, it is well known that VAEs often converge to local minima in which the latent variables are not used, a phenomenon known as posterior collapse \cite{LucasTGN19}. This can be further exacerbated in hierarchical models as there may be less incentive for higher level latent variables to be used \cite{ZhaoSE17}. To tackle this issue we choose to formulate Eq.\ \ref{eq:order_l_loss} as a constrained optimization problem using the Generalized ELBO with Constrained Optimization (GECO) method \cite{Rezende2018GeneralizedEW}. The basic idea is to formulate the optimization of the ELBO as a minimization of the KL term subject to the constraint that the average reconstruction error term is within a chosen value. Using GECO showed consistently better use of the latent variables and the conditioning information (which is key to learning useful belief representations), as opposed to the standard VAE objective. We give the details of the GECO hyper-parameters used for the experiments in Appendix \ref{sec:appendix_agent_architectures}.

\section{Running with Scissors game.}
\label{sec:appendix_games}

\begin{figure*}[htb]
     \centering
     \begin{subfigure}[b]{0.65\textwidth}
         \centering
         \includegraphics[width=\textwidth]{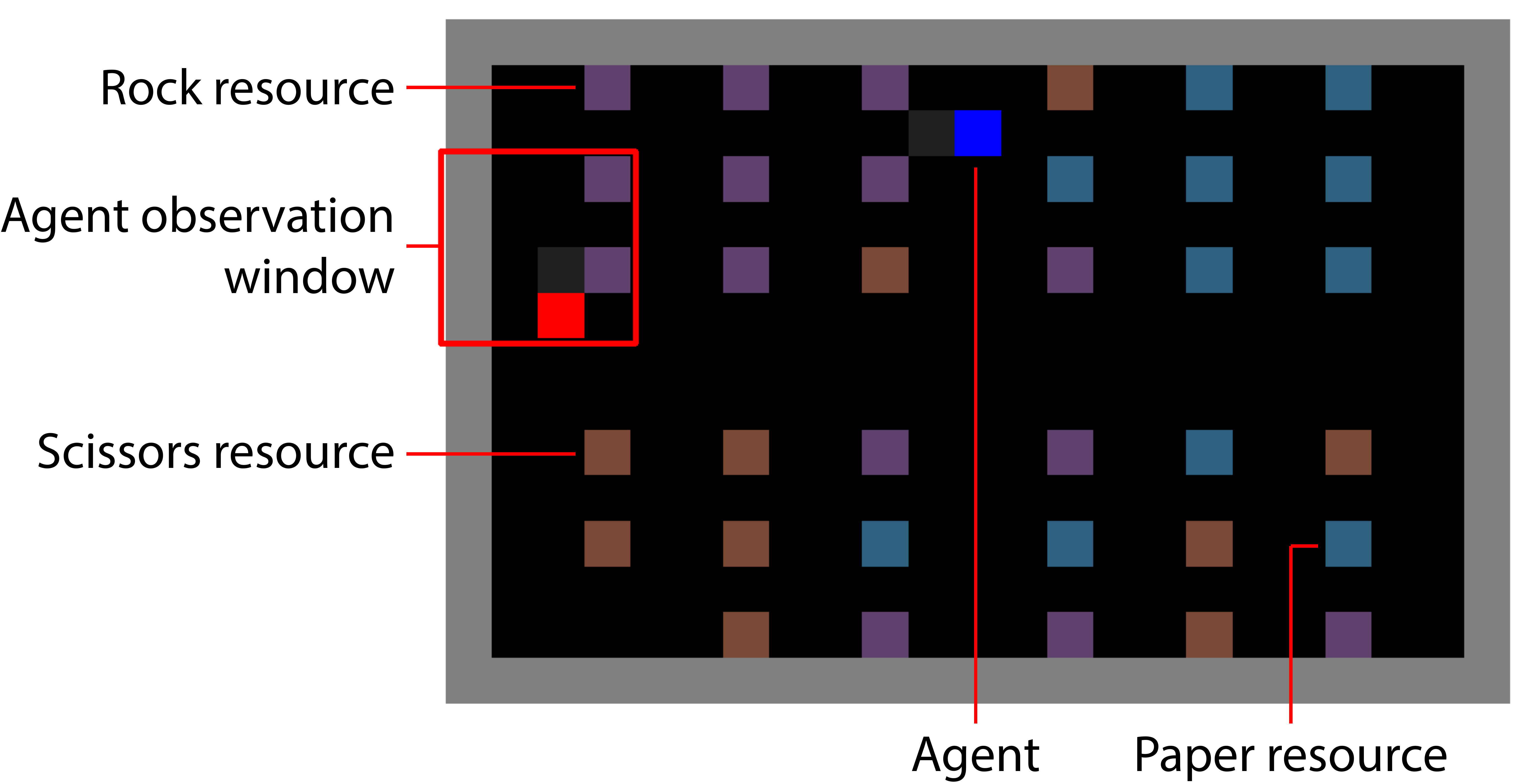}
         \caption{Running with Scissors.}
         \label{fig:rws_gameplay}
     \end{subfigure}
        \caption{Competitive and mixed-motive partially observable Markov game.}
        \label{fig:markov_games}
\end{figure*}

In Running with Scissors, when a player steps on a tile with a resource, it picks the resource up. The resource is subsequently removed from the environment and allocated into the player's inventory $v$. Player inventories are initialized to contain one of each resource type. Resources are placed as shown in Fig.\ \ref{fig:rws_gameplay} with four main piles (rock, paper, scissors and a random pile). The resources initialized in every episode are fixed except for the fourth pile, which randomizes nine resources. Players have a special ``tagging'' action which can end an episode early. If a player uses their tagging action when their opponent is in a small area in front of them, the episode terminates. Once the episode ends, player rewards are calculated according to a standard anti-symmetric formula for rock-paper-scissors: $r^0=v^0M(v^1)^\top$ (see \cite{hofbauer2003evolutionary, vezhnevets2019options}). To maximize reward, players should correctly identify which resource the opponent is collecting and then collect the correct counter-resource. Finally, each player receives RGB visual observations of a $4\times4$ RGB window around their position, with a slight offset. Players also observe their current inventory of rocks, papers and scissors.

\section{Tiger game with second order beliefs.}
\label{sec:appendix_tiger_b2}

\begin{figure}[htb]
    \centering
    \includegraphics[width=0.75\textwidth]{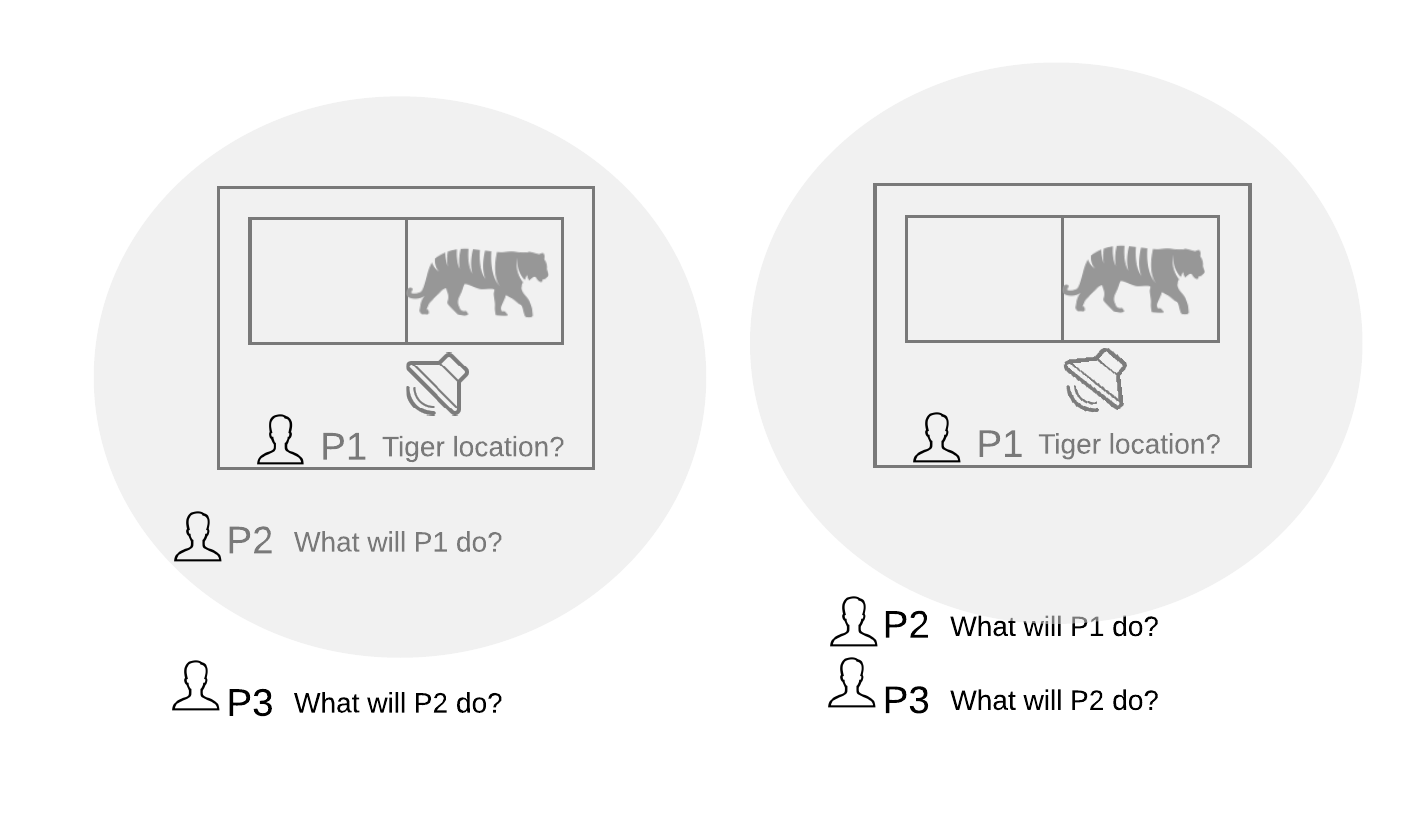}
    \caption{Second version of the Tiger game which illustrates two possible scenarios. Gray circle indicates the range of the tiger's growl. The other two initial scenarios not represented here are the corresponding cases when the tiger is in the left door.}
    \label{fig:tiger_env_b2}
\end{figure}

We now describe a version of the Tiger game with a third player. The game goes as described above, with P1 having three possible actions $\{OL, OR, L\}$ for each of the 10 rounds. However, at the beginning of the episode, P2 is placed either close to the room where P1 is, or far. Figure \ref{fig:tiger_env_b2} reflects these two situations.
When placed close, P2 can hear the growl (but not where the tiger is exactly) as in the original version, but when placed far, P2 cannot hear the growl through the whole episode. P2 is aware of its distance, thus their model of P1's distribution of the location of the tiger $S_t$ has to change accordingly. In this game P2 has three possible actions: (1) predict that P1 will choose any door, (2) predict that P1 will wait and listen, or (3) decide not to commit to any prediction (`wait'). P2 gets negatively rewarded when committing to the wrong prediction so it is encouraged to `wait' when unsure of P1's knowledge, i.e.\ when P2 is placed far.
The third player P3 is always placed far from the room. This player never hears the tiger, but observes whether P2 is close or far from the room. Based on this observation, at every time step P3 must predict whether P2 will either choose to `wait', or not (i.e.\ if P2 commits to a prediction of P1). 

To show how a single sample from P2's $B_t^1(P2)$ does not suffice as a target for our belief model (at any time-step), first consider the optimal nested samples of P2 using $K=4$ samples. We know there are three possible $B^0(P1)$ models, and we'll refer to them as: (1) P1 knows the tiger is left $B_{TL}^0$, (2) P1 knows tiger is right $B_{TR}^0$, (3) P2 is unsure $B_{U}^0$. Table \ref{tab:tiger_scenarios} shows examples of possible samples at the first order given the four different scenarios. As we can see, by providing multiple samples, P3 knows that if P2 is close to the room, there's a chance that P2 is confident that P1 is unsure and that there is an increasing probability at every round that there might have been a growl (because P2 must have heard it). Conversely, if P2 is far, P3 knows that P2 will never be fully certain that P1 knows where the tiger is, so it can predict that P2 will not commit to a prediction. Below we show experimental results on the ability of the policy to use the model to solve the corresponding tasks for both Tiger games.

\begin{table}[htb]
    \centering
    \caption{Nested samples of $B^1(P2)$ for the Tiger game, here we show 4 representative samples from B1. In bold, we highlight the first sample in the collection to emphasize that the distribution over that one sample does not allow distinguishing between the event of P2 being close from P2 being far}
    \label{tab:tiger_scenarios}
    \begin{tabular}{lll}
        \toprule
        {} & P2 is close & P2 is far \\
        \midrule
        Growl & (\bm{$B_{TL}^0$}, $B_{TR}^0$, $B_{TL}^0$, $B_{TR}^0$)  & (\bm{$B_{TL}^0$}, $B_{U}^0$, $B_{U}^0$, $B_{TR}^0$)  \\
        No growl & (\bm{$B_{U}^0$}, $B_{U}^0$, $B_{U}^0$, $B_{U}^0$)  & (\bm{$B_{U}^0$}, $B_{TL}^0$, $B_{U}^0$, $B_{TR}^0$)  \\
        \bottomrule
    \end{tabular}
\end{table}

\section{Implementation details.}
\label{sec:appendix_agent_architectures}

\subsection{Core architecture.}

The architecture used for our agents with belief models is composed of fairly standard components used for deep nets. For Running with Scissors, the observation encoder in Fig.\ \ref{fig:architecture} is composed of one convolutional network with one layer (16 feature maps, strides=1, kernel size=3x3), connected with an MLP with three layers of [128, 64, 64] hidden units respectively. The nonlinear activations are ReLu for all the networks. Both RNNs that produce $f_t$ and $\hat{b}_t$ are GRUs; $\hat{b}_t$ has 128 units.
The policy takes as input the concatenated internal representations of $[f^{mf}_t, \hat{b_t}]$, and computes the actions with a two layer MLP of [64, 64] hidden units respectively followed by a linear projection that outputs the policy logits. The value network follows the same architecture, but the linear layer predicts the value function instead.

\subsection[Model architecture for B0]{Model architecture for $B_0$.}

The belief model $B_0$ for an agent $i$ at time $t$, $B^0_t(i)$, models the state $S_t$ using an autoregressive model given by
\begin{equation}
   B^0_t(i) = \prod_d^{D_{\text{env}}} p_{\phi}(S_{d,t}|\hat b_t)p_{\phi}(S_{d,t}|S_{<d,t}, \hat b_t(i)),
\end{equation} 
where $D_{\text{env}}$ is the dimensionality of the state. We use the powerful and general-purpose autoregressive model MADE \cite{GermainGML15}. We implement this model with 4 residual layers (following the ResMADE architecture proposed in \cite{DurkanN19}), where each layer is composed of 128 hidden units. The conditioning information $\hat b_t$ is incorporated by first computing  a linear projection of $\hat b_t$, and then adding that projection both additively and multiplicatively to the hidden activations in each of the residual layers. That is, the hidden units of each residual layer $h$ are transformed $h' = h \cdot p + p$, where $\cdot$ is an element-wise product, and $p = W \hat b_t$ is the learned linear projection.

\subsection[Model architecture for B1]{Model architecture for $B_l$.}
\label{sec:appendix_b_l}
For these experiments, we model the latent variables as diagonal Gaussian variables, where the mean and variances are learned as a function of the conditioning variables. We use a flexible conditional prior for the generative model using the ResMADE architecture mentioned above, with four residual layers, where each layer is composed of 128 hidden units. The conditioning variables $\hat b_t$ are concatenated to the model's hidden units in the same fashion as described for the state model of $B_0$. Note that for intermediate latent variables, i.e.\ $p(Z^{l-1}_t|z^{l-1}, \hat b_t)$ the conditioning is applied to the concatenation of $z^{l-1}$ and $\hat b_t$. It is important to note that the hierarchical prior used should be efficient at evaluating log-likelihoods as well as sampling. To improve efficiency in sampling, we choose the autoregressive structure of ResMADE to be applied only at the level of the number of players. That is, for three players, from the perspective of player 1, we have $p(Z^l(2), Z^l(3)|\hat b^{l}(1)) = p(Z^l(2)|\hat b^{l}(1))p(Z^1(3)|Z^1(2), \hat b^1(1))$, and each conditional models the latent vector as diagonal Gaussian as described. Below we describe the architecture used to encode the nested representation of one player.

We compute the proposal $q(Z^1|b^{1}, \hat b)$ using an order invariant function over the samples $b^{1}$: first each sample $k$ is processed using an MLP of two layers of [64, 64,] units respectively, then the encodings for every sample $e_k$ are all summed to form the aggregated embedding $m = \sum^K_k e_k$. The conditioning neural code $\hat b_t$ is projected to a linear function (same output size as $m$) W, and is combined in two ways. First, it is element-wise multiplied with $m$ after squashing it via a sigmoid, and then concatenated with $\hat b$: $m' = [m \cdot \sigma(p), p]$, where $p = W \hat b_t$. The concatenated output $m'$ is then processed with an MLP (2 layers of [64, 64] units), followed by a linear mapping to output the parameters of the posterior distribution given by a diagonal Gaussian $\bf \mu$ and $\bf \sigma$. The number of latent variables used for the experiments in the different environments is specified in the section below.

\subsection{Model architecture for trajectory-prediction agents.}

\label{sec:appendix_trajectory_agent}

This baseline is based on learning its belief representation by predicting observations $Y_t$ and actions $A_t$ of the opponents, instead of the state $S_t$ as does $B^0$. We use the same autoregressive distribution used for the $B^0$ model however to separately predict the observation and actions (i.e.\ 4 residual blocks of 128 units each). The observations are discretized to  $4\times 4 \times 255$ for Running with Scissors. The outputs of the ResMADE model are thus the logits of the Categorical distribution for each observation conditional.

\subsection{Experiment hyper-parameters.}

For the purpose of these experiments, the following hyper-parameters showed to work well across all types of agents in early experiments.

For Running with Scissors:
\begin{itemize}
\item Learning rate is 0.0003.
\item Entropy cost weight is 0.015.
\item Model-free GRU $f^{mf}_{\theta_t}$ state sizes is 128, and belief GRU is ($f^{b}_{\theta_t}$) size is 128.
\item For B1, we use a 64 diagonal Gaussian dimensional latent variable for each $n-1$ players.
\item Belief of order 0 loss weight = 10. Belief of order 1 loss weight = 1. 
\item GECO parameters: reconstruction loss threshold of 1. Initial Langrangian multiplier value of 1, and range of feasible Lagrangian multiplier values between (0.1, 40).

\end{itemize}

For Tiger experiments, the hyper-parameters that are different are::

\begin{itemize}
\item Learning rate is 0.0002.
\item Belief of order 0 weight = 100, Belief of order 1 weight = 10.  Belief of order 2 weight = 10. 
\item For for belief of order 1 and 2, we use an 8 dimensional diagonal Gaussian latent variable.
\item GECO parameters: reconstruction loss threshold of 0.25. Initial Langrangian multiplier value of 1, and range of feasible Lagrangian multiplier between (0.1, 40). For belief model of order 2, we encourage the latent variables to be used by adding a additional constraints: $1 < \expec_{\tau_t} \text{KL}^{l=2}_t < 5$. That is, we encourage a committed rate of KL divergence (wrt. posterior and prior of the latent variables of second order) to be in the a desired range.

\end{itemize}

Finally, the gradients from the policy and belief losses are applied using the RMSProp optimizer with momentum of 0.0 and decay of 0.99 for all environments.

\end{document}